\documentclass{ecai}
\usepackage{times}
\usepackage{graphicx}
\usepackage{latexsym}
\usepackage{multirow}
\usepackage{amssymb}
\usepackage{multirow}
\usepackage{lipsum}
 
\begin{document}

\title{Crowd Counting via Hierarchical Scale \\
Recalibration Network}

\author{Zhikang Zou\textsuperscript{\textmd{1*}} \and Yifan Liu\textsuperscript{\textmd{1*}} \and Shuangjie Xu\textsuperscript{\textmd{2}} \and Wei Wei \textsuperscript{\textmd{1}} \\ \and Shiping Wen \textsuperscript{\textmd{3}} \and Pan Zhou \textsuperscript{\textmd{1}$\dagger$}}


\maketitle
\bibliographystyle{ecai}

\newcommand\blfootnote[1]{%
\begingroup
\renewcommand\thefootnote{}\footnote{#1}%
\addtocounter{footnote}{-1}%
\endgroup
}

\begin{abstract}
    The task of crowd counting is extremely challenging due to complicated difficulties, especially the huge variation in vision scale. Previous works tend to adopt a naive concatenation of multi-scale information to tackle it, while the scale shifts between the feature maps are ignored. In this paper, we propose a novel Hierarchical Scale Recalibration Network (HSRNet), which addresses the above issues by modeling rich contextual dependencies and recalibrating multiple scale-associated information. Specifically, a Scale Focus Module (SFM) first integrates global context into local features by modeling the semantic inter-dependencies along channel and spatial dimensions sequentially. 
    In order to reallocate channel-wise feature responses, a Scale Recalibration Module (SRM) adopts a step-by-step fusion to generate final density maps. Furthermore, we propose a novel Scale Consistency loss to constrain that the scale-associated outputs are coherent with groundtruth of different scales. With the proposed modules, our approach can ignore various noises selectively and focus on appropriate crowd scales automatically. Extensive experiments on crowd counting datasets (ShanghaiTech, MALL, WorldEXPO’10, and UCSD) show that our HSRNet can deliver superior results over all state-of-the-art approaches. More remarkably, we extend experiments on an extra vehicle dataset
    , whose results indicate that the proposed model is generalized to other applications.
\end{abstract}

\footnote{Huazhong University of Science and Technology, China. Email: zhikangzou001@gmail.com, \{u201712105,weiw,panzhou\}@hust.edu.cn}
\footnote{Deeproute.ai,China. Email:shuangjiexu@deeproute.ai}
\footnote{University of Electronic Science and Technology, China. Email: wenshiping@uestc.edu.cn \\ \noindent\textsuperscript{$*$} Equal contributions. \hspace{0.2cm} \textbf{$\dagger$} Corresponding author}


\section{Introduction}
The task of crowd counting aims to figure out the quantity of the pedestrians in images or videos. It has drawn much attention recently due to its broad possibilities of applications in video surveillance, traffic control and metropolis safety. What's more, the methods proposed for crowd counting can be generalized to similar tasks in other domains, including estimating the number of cells in a microscopic image \cite{lempitsky2010learning}, vehicle estimation in a traffic congestion situation \cite{guerrero2015extremely} and extensive environmental investigation \cite{french2015convolutional}.

\begin{figure}[t]
\centerline{\includegraphics[width=\linewidth]{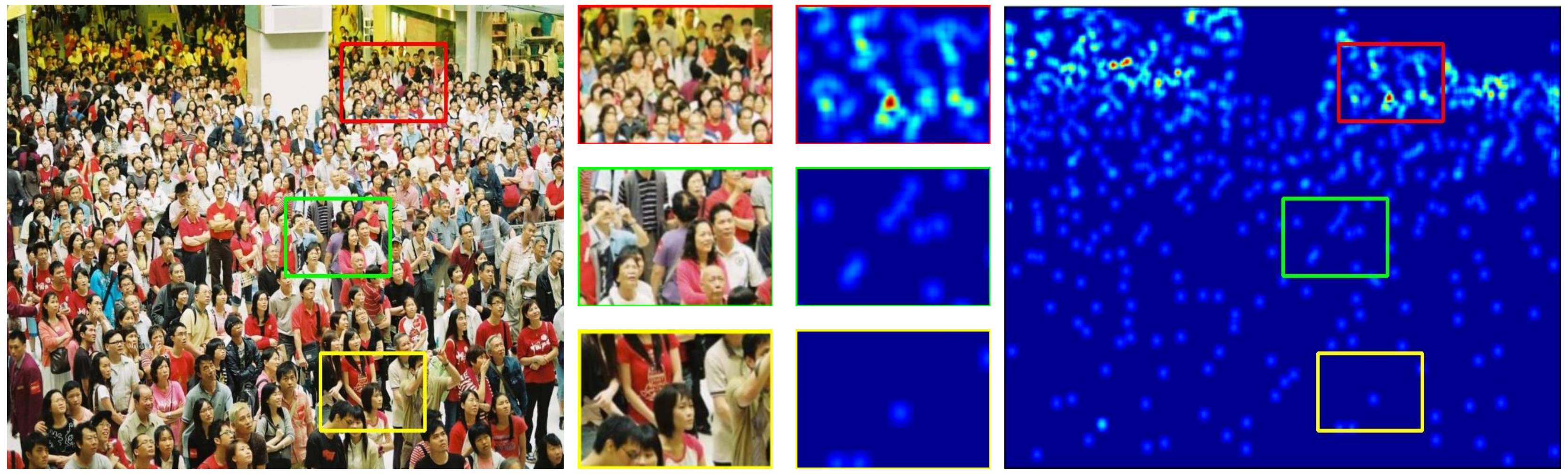}}
\caption{Regions of certain sizes exhibit diverse pedestrian distribution among different locations of images. This shift also exists across different images, which proves the huge scale variations in crowd counting.} \label{abstract}
\end{figure}

With the rapid growth of convolutional neural networks, many CNN-based methods \cite{liu2018leveraging,babu2018divide,wang2019learning} have sprung up in fields of crowd counting and have made promising progress. However, dealing with the large density variations is still a difficult but attractive issue. As illustrated in Figure \ref{abstract}, the crowd density of certain sizes varies in different locations of the images. Such a density shift also exists in the patches of the same sizes across different images. To address the scale variation, substantial progress has been achieved by designing multi-column architectures \cite{sindagi2017generating,sindagi2017cnn}, adaptively fusing features pyramid \cite{kang2018crowd}, and modifying the receptive fields of CNNs \cite{li2018csrnet}. Although these methods alleviate the scale problem to some extent, they suffer from two inherent algorithmic drawbacks. On the one hand, each sub-network or each layer in these models treats every pixel of the input equally while ignoring their superior particularity on the corresponding crowd scales, thus the noises will be propagated through the pipeline flow. On the other hand, directly adding or concatenating multi-scale features causes the scale chaos since each feature map contains abundant scale shifts of different degrees.

To settle the above issues, we propose a novel Hierarchical Scale Recalibration Network (HSRNet) to leverage rich contextual dependencies and aggregate multiple scale-associated information. Our training phase contains two stages: the Scale Focus Module (SFM) and the Scale Recalibration Module (SRM). Since the receptive field sizes of the sequential convolutional layers in a deep network are increasing from shallow to deep, the scales of pedestrians they can capture are different from each other. This can deduce two inferences: 1) the deeper the network flows, the wider the scale range can be captured by the corresponding convolutional layers, 2) sensitivity to different scales varies across different layers of the network. Thus, we connect a Scale Focus Module (SFM) to each convolutional layer in the backbone network, which integrates global context into local features to boost the capability of intermediate features on the corresponding scales. More specifically, SFM firstly compresses the input features in the spatial dimension and generates a set of channel-wise focus weights, which are utilized to update each channel map. Thus, each layer can emphasize the matching scales degree by adjusting channel-wise feature responses adaptively. Similarly, the context along the channel axis in the feature map is squeezed to generate a spatial-wise focus mask and it is applied to update features at all positions using element-wise multiplication. Note that this strategy enhances that the output features focus more on the patches of images with appropriate scales instead of treating every pixel equally. By incorporating this module in the network, intermediate layers can focus on 'which' scale degree and 'where' scale distributes simultaneously and hence enhance the discriminant power of the feature representations.

In a hierarchical architecture, the scale space is increasing from shallow to deep, which means feature maps from different layers contain scale asymmetry. Due to this, naive average or concatenation of multiple features is not an optimal solution. We propose a novel Scale Recalibration Module (SRM) to further achieve adaptive scale recalibration and generate multi-scale predictions at different stages. Specifically, this module takes the feature maps processed by the SFM as input and then slice these features in channel dimension. Since each channel is associated with a certain scale, the pieces represented by the corresponding scales can be recombined through stacking to obtain scale-associated outputs. In this case, each output can capture a certain scale of crowds and give an accurate prediction on the patches of that scale. We fuse these outputs to generate the final density map, which could have accurate responses on crowd images of diverse scales.
To enforce the network produces consistent multi-scale density maps, we propose a Scale Consistency loss to pose supervision on scale-associated outputs. It is computed by generating multi-scale groundtruth density maps and optimizing each side output towards the corresponding scale maps. 

In general, the contributions of our work are three-folds:
\begin{itemize}
\item We propose a Scale Focus Module (SFM) to enhance the representation power of local features. By modeling rich contextual dependencies among channel and spatial dimensions, different layers in the network can focus on the appropriate scales of pedestrians.

\item We propose a Scale Recalibration Module (SRM) to recalibrate and aggregate multi-scale features from sequential layers at different stages. It significantly enhances the adaption ability of the structure to the complicated scenes with diverse scale variations.

\item We propose a Scale Consistency loss to supervise the scale-associated outputs at different scale level, which enforces the network produces consistent density maps with multiple scales.
\end{itemize}

\section{Related Work}
The previous frameworks are mainly composed of two paradigms: 1) people detection or tracking \cite{wu2005detection}, 2) feature-based regression \cite{chen2012feature}. However, these methods are generally unpractical due to their poor performance and high computation. As the utilization of Convolutional Neural Network (CNN) has boosted improvements in various computer vision tasks \cite{xiong2016person,zou2019attend,bartos2016learning,liu2018leveraging}, most recent works are inclined to use CNN-based methods. They tend to generate accurate density maps whose integral indicates the total number of crowds. However, it is still challenging to achieve precise pedestrians counting in extremely complicated scenes for the presence of various complexities, especially scale variations.

To tackle the above issues, many existing approaches focus on improving the scale variance of features using multi-column structures for crowd counting \cite{7780439,sindagi2017generating,sindagi2017cnn,cheng2019improving}. Specifically, they utilize multiple branches, each of which has its own filter size, to strengthen the ability of learning density variations across diverse feature resolutions. Despite the promising results, these methods are limited by two drawbacks: 1) a large amount of parameters usually results in difficulties for training, 2) the existence of ineffective branches leads to structure redundancy.

In addition to multi-column networks, some methods adopt multi-scale but single-column architecture \cite{8497050,zou2019enhanced}. For instance, Zhang \textit{et al}. \cite{zhang2018crowd} propose an architecture which extracts and fuses feature maps from different layers to generate high-quality density maps. In SANet \cite{cao2018scale}, Cao \textit{et al}. deploy an encoder-decoder architecture, in which the encoder part utilizes scale aggregation modules to extract multi-scale features and the decoder part generates high-resolution density maps via transposed convolutions. Li \textit{et al}. \cite{li2018csrnet} replace pooling layers with dilated kernels to deliver larger receptive fields, which effectively capture more spatial information. 
After this, Kang \textit{et al}. \cite{kang2018beyond} design two different networks and evaluate the quality of generated density maps on three crowd analysis tasks.

However, all these methods directly fuse multi-layer or multi-column features to generate the final density maps. It ignores the unique perception of each part to the scale diversity and thus causes the scale chaos in the output result. Besides, attention-based methods \cite{8578843,woo2018cbam} have proved their effectiveness in several deep learning tasks. These approaches work by allocating computational resources towards the most relevant part of information. In this paper, we propose a novel Hierarchical Scale Recalibration Network (HSRNet) to resolve the severe difficulties of scale variations. Our method differs in two aspects: firstly, instead of treating the whole images, our method is able to focus on the appropriate scale of the crowds; secondly, the scale recalibration takes place to effectively exploit the specialization of the components in the whole architecture.

\begin{figure*}[t]
\centerline{\includegraphics[width=\linewidth]{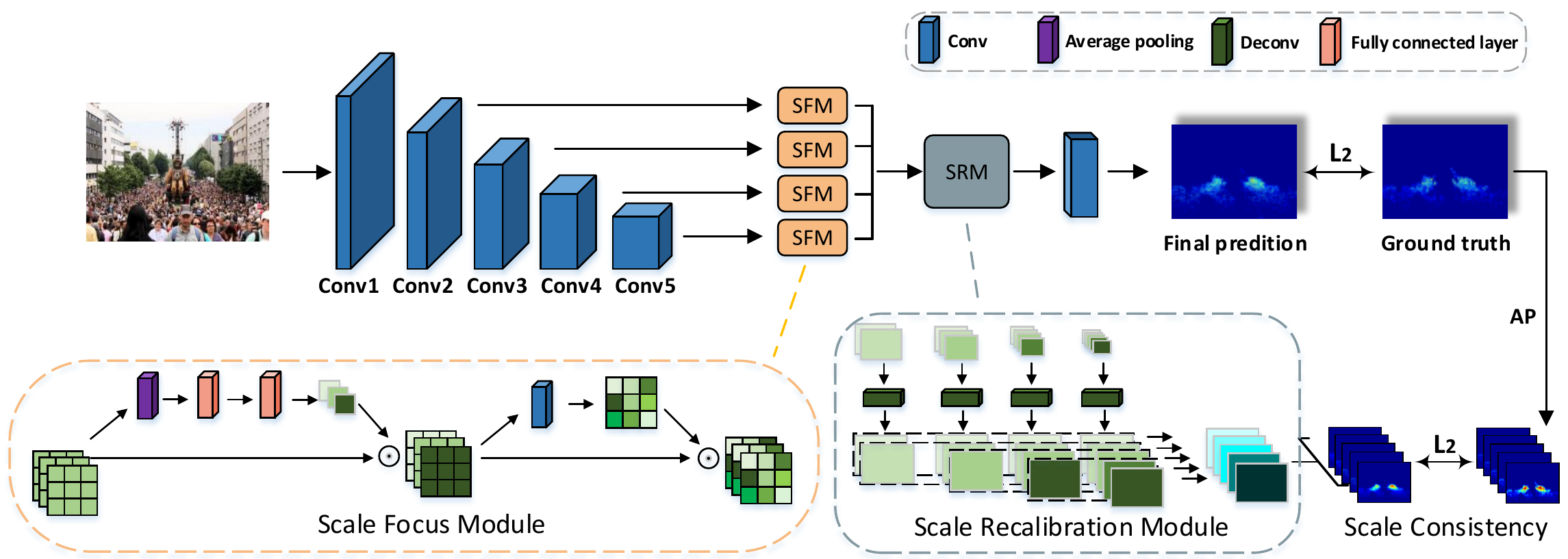}}
\caption{The detailed structure of the proposed Hierarchical Scale Recalibration Network (HSRNet). The single column baseline network on the top is VGG-16, at the end of each convolutional layer, feature maps are sent to Scale Focus Module (SFM) to get refined feature maps. Then, the Scale Recalibration Module (SRM) process these features to get multi-scale predictions, which are finally fused by a 1x1 convolution to generate the final density map.} \label{whole}
\end{figure*}

\section{Our Approach}
The primary objective of our model is to learn a mapping $F:X \to Y$, in which $X$ means input image data, and the learning object $Y$ has two choices: density map or total crowd counting number. Motivated by the aforementioned observations, we choose the density map as the main task of our model in the training phase to involve the spatial distribution information for a better representation of crowds, which is realizable with a network of fully convolutional structure. In this paper, we propose a novel proposed Hierarchical Scale Recalibration Network (HSRNet) to address the scale variations in crowd counting, the overall architecture of which is shown in Figure \ref{whole}. For fair comparison with previous works \cite{sam2017switching, li2018csrnet,bai2019crowd}, we choose VGG-16 \cite{simonyan2014very} network as the backbone by reason of its strong representative ability and adjustable structure for subsequent feature fusion. The last pooling layer and the classification part composed of fully-connected layers are removed for the task of counting requires pixel-level predictions and preventing spatial information loss. Thereby our backbone consists of five stages (Conv1 $\sim$ Conv5 respectively). To focus on the appropriate scales of pedestrians, we connect the proposed Scale Focus Module (SFM) to the last convolutional layer in each stage except for the first one to obtain the fine-grained features from multiple layers. The reason why we carve the first stage is the receptive field sizes of the first convolutional layers are too small to obtain any information of crowds. Since the scales of pedestrians the convolutional layers can capture varies across different stages, we send these features after the process of SFM to the Scale Recalibration Module (SRM) to reallocate scale-aware responses by a slice/stack strategy. Thus each side output corresponds to a certain scale and provides an accurate crowd prediction of that scale. With the utilization of deconvolutional layer, each prediction stays the same resolution as the input image. The final density map can be generated by fusing these scale-associated outputs. To guarantee that each scale-associated output is optimized towards a specific direction, we propose a Scale Consistency loss to supervise the target learning.

\subsection{Scale Focus Module}
The Scale Focus Module is designed to enforce sequential layers at different stages focus on the appropriate scales of pedestrians by encoding the global contextual information into local features. Since the receptive fields of each convolutional layer are accumulating from shallow to deep, the scale space they are able to cope with increases accordingly. Besides, there are specific shifts between their representation ability on crowd scales, which indicates that different stages should be responsible for the corresponding scales. With these observations, we generate adaptive weights to emphasize feature responses in the channel and spatial dimension respectively. 

In channel dimension, each channel map of immediate features can be regarded as a scale-aware response. For a high-level feature, different channels are associated with each other in terms of semantics information. By exploring the inter-channel interdependencies, we could generate channel-focus weights to modify the ratio of different channels with corresponding scales. Similar strategy for spatial dimension, treating the whole images equally is improper since the various crowd distribution leads to different scale space in local patches. However, local features generated merely by standard convolutions are not able to express the whole semantic information. Thus, we generate spatial-focus weights to select attentive regions in the feature maps, which enhances the representative capability of features. Since the channel-focus and spatial-focus weights attend to 'which' scale degree and 'where' scale distributes, they are complementary to each other and the combination of them could boost the discriminant power of the feature representations.

Formally, given an image \textit{X} of size $3 \times H \times W$, the output features from the Conv\textit{(i+1)} layer of the backbone are defined as $F = \{F_i, i=1,2,3,4\}$. Due to the existence of pooling layers, $F_i$ has a resolution of $\frac{H}{2^{i}} \times \frac{W}{2^{i}}$. For the output feature $F_i \in R ^ {C_i \times \frac{H}{2^{i}} \times \frac{W}{2^{i}}}$, we first squeeze the global spatial information to generate channel-wise statics $Z_i \in R^{C_i}$ by utilizing global average pooling function $H_{avg}$. Thus the $j$-th channel of $Z_i$ is defined as
\begin{equation}
    Z_{ij} = H_{avg}(F_{ij}) = \frac{1}{\frac{H}{2^{i}} \times \frac{W}{2^{i}}}\sum^{\frac{H}{2^{i}}}_{m=1}\sum^{\frac{W}{2^{i}}}_{n=1}F_{ij} \left(m,n\right)
\end{equation}
where $F_{ij} \left(m,n\right)$ represents the pixel value at position $(m,n)$ of the  $j$-th channel of $F_i$. Such channel statistic merely collects local spatial information and views each channel independently, which fails to express the global context. Therefore, we add fully-connected layers and introduce a gating mechanism to further capture channel-wise dependencies. The gating mechanism is supposed to meet with two criteria: first, it should be capable to exploit a nonlinear interaction among channels; second, to emphasize multiple channels, it should capture a non-mutually-exclusive relationship. We use the sigmoid activation to realize the gating mechanism:
\begin{equation}
    S_i=Sigmoid\left(W_{2} \cdot ReLU\left(W_{1}Z_i\right)\right)\label{ratio_eq}
\end{equation}
where $W_1 \in R^{C_i/r \times C_i}$, $W_2 \in R^{C_i \times C_i/r}$. This operation can be parameterized as two fully-connected (FC) layers, with one defines channel-reduction layer (reduction radio $r$ = 64) and the other represents channel-increasing layer.
After this non-linear activation, we combine the channel-focus weights $S_i$ and the input feature $F_i$ using the element-wise multiplication operation to generate the immediate feature $\hat{S}_i$.
\begin{equation}
    \hat{S}_i = S_i \cdot F_i,  i \in \{1,2,3,4\} 
\end{equation}
Thus, the global channel information is encoded into local features. Then we take the mean value of $\hat{S}_i$ among channels to generate spatial statistic $M_i \in R^{1 \times \frac{H}{2^{i}} \times \frac{W}{2^{i}}}$. After squeezing the information among channels, we feed the spatial statistic $M_i$ into a convolutional layer to generate a spatial-focus weight $\hat{M_i}$:
\begin{equation}
    M_i = \frac{1}{C_i} \sum\nolimits_{i=1}^{C_i} {\hat{S}_i}, \quad \hat{M_i} = Sigmoid(H_c(M_i))
\end{equation}
where $H_c$ indicates the convolution process. Here, the kernel size of the convolutional layer is set to 7, which is capable of providing a broader view. Then we perform an element-wise multiplication operation between $\hat{M_i}$ and $\hat{S}_i$ to obtain the final output $\hat{F_i}$:
\begin{equation}
    \hat{F}_i = \hat{M_i} \cdot \hat{S}_i, i \in \{1,2,3,4\}
\end{equation}
Noted that the spatial-focus weight is copied to apply on each channel of the input in the same way.

\subsection{Scale Recalibration Module}
With the network getting deeper, deep layers can capture more complex and high-level features while shallow layers can reserve more spatial information. Therefore, by fusing features from low-level layers with those from high-level layers, our network can extract stable features no matter how complicated the crowd scenes are. Unlike previous works \cite{sindagi2017generating,kang2018crowd}, We design a scale recalibration module (SRM) to recalibrate and aggregate multi-scale features rather than direct average or concatenation. 

Based on the above analysis, the deep layer has a wider range of scale space and meanwhile has a stronger response on the larger scale of crowds. Formally, assuming that the outputs of the Scale Focus Module are $\hat{F}=\{\hat{F_i},i=1,2,3,4\}$, we first send these features into a $1 \times 1$ convolutional layer and then a deconvolutional layer respectively to obtain multi-scale score maps $E_{i}$, 
\begin{equation}
    E_{i+1} = H_{dc}(H_c(\hat{F}_i))
\end{equation}
where 
$H_{dc}$ is the deconvolution operation. Here, the channel numbers of the sequential layers are $2, 3, 4, 5$ from $E_2$ to $E_5$ respectively, which corresponds to the scale space contained in each stage. The obtained multi-scale score maps $E_{i}$ contain multi-scale information from layers of different depths. However, the information is chaotic. For instance, $E_{5}$ captures multi-scale information delivered from Conv\_\{1,2,3,4,5\} since low-level features are transmitted to the latter stages in the backbone. To recalibrate channel-level statistics, we adopt a slice/stack strategy. Specifically, we slice each score map into piece along its channel dimension to obtain a feature map $E_{ij}$ where $j$ means the channel number, and then group them into five multi-scale features sets from $E_2$ to $E_5$: \{$E_{21}$,$E_{31}$,$E_{41}$,$E_{51}$\},
\{$E_{22}$,$E_{32}$,$E_{42}$,$E_{52}$\},\{$E_{33}$,$E_{43}$,$E_{53}$\},\{$E_{44}$,$E_{54}$\},\{$E_{55}$\}. Each set is associated with a certain scale and we stacked features in each set respectively to generate the corresponding multi-scale predictions $D = \{D_i, i = 1,2,3,4,5\}$. By utilizing this strategy, each prediction is able to provide an accurate number of pedestrians on a certain scale. These predictions are complementary to each other and the combination of them will cover the crowd distributions with various scale variations. Thus we send them into a $1 \times 1$ convolutional layer to generate the final density map $D_0$. Overall, the scale-specific prediction is obtained only with convolutional layer and slice/stack strategy, which is parameter-saving and time-efficiency. With the Scale Recalibration Module, the final output is robust to diverse crowd scales in highly complicated scenes.
 
\subsection{Scale Consistency loss}
The groundtruth density map $D^{GT}$ can be converted from the dot maps which contain the labeled location at the center  of the pedestrian head. Suppose a pedestrian head at a pixel $x_i$, we represent each head annotation of the image as a delta function $\delta(x-x_{i})$ and blur it with Gaussian kernel $G_{\sigma}$ ($\sigma$ refers to the standard deviation). So that the density map $D^{GT}$ is obtained via the formula below:
    \begin{equation}
        D^{GT}(x) =\sum_{i \in S}\delta(x-x_{i})*G_{\sigma_{i}},\ {\rm with}\ \sigma_{i}=\beta\bar{d_{i}} \label{adaptive}
    \end{equation}
Where $S$ is the amount of head annotations, $\bar{d_{i}}$ refers to the average distance among $x_{i}$ and its $k$ nearest annotations and $\beta$ is a parameter. We use this geometry-adaptive kernels following MCNN \cite{7780439} to tackle the perspective distortion in highly complicated scenes. The Euclidean distance is utilized to define the density map loss, which can be formulated as follows:
    \begin{equation}
       L(\theta)=\frac{1}{2N}\sum_{i=1}^{N}||D(X_{i};\theta)-D_{i}^{GT}||_{2}^{2} 
    \end{equation}
where $\theta$ denotes the parameters of the network and $N$ is the amount of image pixels. Usually, this loss is merely calculated between the final density map and the groudtruth map. In this paper, we propose a novel Scale Consistency loss to guide the multi-scale predictions to be optimized towards its corresponding scale map. Specifically, we use the average pooling to obtain the groundtruth pyramid $D^{GT}_i,i=1,2,3,4,5$. The receptive fields of filters are 1, 2, 4, 8, 16, respectively. Then these maps are upsampled to the same size as the original image through a bilinear interpolation. We can compute the loss pairs $\{D_i,D^{GT}_i\}$ and obtain the loss pyramid $\{L_0,L_1,L_2,L_3,L_4,L_5\}$. The total loss of our model can be defined as:
\begin{equation}
    L = L_0 + \lambda_i \cdot L_i, i \in \{1,2,3,4,5\}
\end{equation}
where $\lambda_i$ is a scale-specific weight. It can be gradually optimized and adaptively adjust the ratio between losses.

\section{Experiments}
In this section, we evaluate our method on four publicly available crowd counting datasets: ShanghaiTech, WorldExpo'10, UCSD and MALL. Compared with previous approaches, the proposed HSRNet achieves state-of-the-art performance. Besides, experiments on a vehicle dataset TRANCOS are performed to testify the generalization capability of our model. Furthermore, we conduct throughout ablation studies to verify the effectiveness of each component in our model. Experimental settings and results are detailed below.

\subsection{Implementation Details}
\textbf{Data Augmentation.} We first crop four patches at four quarters of the image without overlapping which is 1/4 size of the original image resolution. By this operation, our training datasets can cover the whole images. Then, we crop 10 patches at random locations of each image with the same size. Also, random scaling is utilized to construct multi-scale image pyramid with scales of 0.8-1.2 incremented in interval of 0.2 times the original image. During test, the whole images are fed into the network rather than cropped patches.

\noindent\textbf{Training Phase.}  We train the proposed HSRNet in an end-to-end manner. The first ten convolutional layers of our model are initialized from the pre-trained VGG-16 \cite{simonyan2014very} while the rest of convolutional layers are initialized by a Gaussian distribution with zero mean a standard deviation of 0.01. We use the Adam optimizer \cite{kingma2014adam} with an initial learning rate of 1e-5. In Eqn. \ref{adaptive}, $k$ is set to 3 and $\beta$ is set to 0.3 following MCNN \cite{7780439}.

\noindent\textbf{Evaluation Metrics.} Following existing state-of-the-art methods \cite{li2018csrnet,sindagi2017cnn}, the mean absolute error (MAE) and the mean squared error (MSE) are used to evaluate the performance on the test dataset, which can be described as follows:
    \begin{equation}
        {\rm MAE}=\frac{1}{N}\sum_{i=1}^{N}|z_{i}-\hat{z_{i}}|, 
    \end{equation}
    \begin{equation}
        {\rm MSE}=\sqrt{\frac{1}{N}\sum_{i=1}^{N}(z_{i}-\hat{z_{i}})^{2}}
    \end{equation}
where $N$ means numbers of image, $z_{i}$ means the total count of the image, and $\hat{z_{i}}$ refers to the total count of corresponding estimated density map.

    \begin{table}
    \renewcommand{\arraystretch}{1.2}
    \caption{Experimental results on ShanghaiTech dataset.}
    \label{shanghai_table}
    \centering
    \begin{tabular}{c|c|c|c|c}
    \hline
     \multirow{2}{*}{Method}&\multicolumn{2}{|c|}{ShanghaiTech Part\underline{\hbox to 0.1cm{}}A} & \multicolumn{2}{|c}{ShanghaiTech Part\underline{\hbox to 0.1cm{}}B}\\
    \cline{2-5}
     &  MAE  &   MSE &   MAE  &   MSE \\
    \hline
    \hline
    MCNN \cite{7780439} & 110.2 & 173.2 & 26.4 & 41.3\\
    \hline
    Switching-CNN \cite{sam2017switching} & 90.4 & 135.0 & 21.6 & 33.4\\
    \hline
    IG-CNN\cite{babu2018divide}& 72.5 & 118.2 & 13.6 & 21.1\\
    \hline
    CSRNet \cite{li2018csrnet}& 68.2 & 115.0 & 10.6 & 16.0\\
    \hline
    SANet \cite{cao2018scale} & 67.0 & 104.5 & 8.4 & 13.6 \\
    \hline
    TEDnet \cite{jiang2019crowd} & 64.2 & 109.1 & 8.2 & 12.8 \\
    \hline
    SFCN \cite{wang2019learning} & 64.8 & 107.5 & 7.6 & 13.0 \\
    \hline
    HSRNet (ours) & \textbf{62.3} & \textbf{100.3} & \textbf{7.2} & \textbf{11.8} \\
    \hline
    \end{tabular}
    \end{table}
    
    \begin{table}
    \renewcommand{\arraystretch}{1.2}
    \caption{Experimental results on UCSD and MALL dataset.}
    \label{ucsd_mall}
    \centering
    \begin{tabular}{c|c|c|c|c}
    \hline
     \multirow{2}{*}{Method}&\multicolumn{2}{|c|}{UCSD dataset} & \multicolumn{2}{|c}{MALL dataset}\\
    \cline{2-5}
     &  MAE  &   MSE &   MAE  &   MSE \\
    \hline
    \hline
    Ridge Regression \cite{chen2012feature} & 2.25 & 7.82 & 3.59 & 19.0 \\
    \hline
    CNN-Boosting \cite{walach2016learning} & 1.10 & - & 2.01 & - \\
    \hline
    MCNN \cite{7780439} & 1.07 & 1.35 & 2.24 & 8.5\\
    \hline
    ConvLSTM-nt \cite{xiong2017spatiotemporal} & 1.73 & 3.52 & 2.53 & 11.12 \\
    \hline
    Bidirectional ConvLSTM \cite{xiong2017spatiotemporal} & 1.13 & 1.43 & 2.10 & 7.6\\
    \hline
    CSRNet \cite{li2018csrnet} & 1.16 & 1.47 & - & - \\
    \hline
    HSRNet (ours) & \textbf{1.03} & \textbf{1.32} & \textbf{1.80} & \textbf{2.28} \\
    \hline
    \end{tabular}
    \end{table}

\subsection{Performance on Comparison}
We evaluate the performance of our model on four benchmark datasets and a vehicle dataset. Overall, the proposed HSRNet achieves the superior results over existing state-of-the-art methods.

\noindent\textbf{ShanghaiTech.} The ShanghaiTech dataset \cite{7780439} consists of 1198 images which contains a total amount of 330165 persons. It is separated into two parts: the one is named as Part A with 482 pictures and the other is Part B with 716 pictures. Part A composed of rather congested images is randomly captured from the web while Part B is comprised of images with relatively low density captured from street views. Following the setup in \cite{7780439}, we use 300 images to form the training set and the left 182 images for testing in Part A, while in Part B, 400 images are used to compose the training set and the remaining 316 are considered as testing set. The comparison results of performance between the proposed HSRNet with some previous methods are reported in Table \ref{shanghai_table}. It is shown that our model achieves superior performance on both parts of the Shanghaitech dataset.

\noindent\textbf{UCSD.} The UCSD dataset\cite{chan2008privacy} contains 2000 frames with size of 158x238 captured by surveillance cameras. This dataset includes relatively sparse crowds with count varying from 11 to 46 per image. Authors also provide the ROI to discard the irrelevant areas in the image. Utilizing the vertexes of ROI given for each scene, we modify the last multi-scale prediction from fusing layers based on the given ROI mask, setting the nerve cells out of the ROI regions not work. We use frames 601 to 1400 as the training set and the rest as testing set. Table \ref{ucsd_mall} illustrates that our model achieves the lowest MAE and MSE compared with previous works, which indicates that HSRNet can perform well in both dense and sparse scenarios.

\begin{table}
\renewcommand{\arraystretch}{1.2}
\caption{Estimation results on WorldExpo'10 dataset.}
\label{world_table}
\centering
\begin{tabular}{c|c|c|c|c|c|c}
\hline
 Method  &  S1 &  S2 &   S3 &  S4 & S5 &  Ave \\
\hline
\hline
MCNN \cite{7780439} & 3.4 & 20.6 & 12.9 & 13.0 & 8.1 & 11.6 \\
\hline
Switching-CNN \cite{sam2017switching} & 4.4 & 15.7 & 10.0 & 11.0 & 5.9 & 9.4 \\
\hline
IG-CNN \cite{babu2018divide}& 2.6 & 16.1 & 10.15 & 20.2 & 7.6 & 11.3 \\
\hline
CSRNet \cite{li2018csrnet}& 2.9 & 11.5 & \textbf{8.6} & 16.6 & 3.4 & 8.6 \\
\hline
SANet \cite{cao2018scale} & 2.6	& 13.2	& 9.0 &	13.3 &	3.0	& 8.2 \\
\hline
TEDnet \cite{jiang2019crowd} & 2.3 & 10.1 & 11.3 & 13.8 & \textbf{2.6} &8.0 \\
\hline
ANF \cite{zhang2019attentional} & \textbf{2.1} & 10.6 & 15.1 & 9.6 & 3.1 & 8.1 \\
\hline
HSRNet (ours) & 2.3 & \textbf{9.6} & 12.7 & \textbf{9.4} & 3.2 & \textbf{7.44} \\
\hline
\end{tabular}
\end{table}

\noindent\textbf{MALL.} The Mall dataset \cite{chen2012feature} is collected from a publicly accessible surveillance web camera. The dataset contains 2,000 annotated frames of static and moving pedestrians with more challenging illumination conditions and severer perspective distortion compared to the UCSD dataset. The ROI and perspective map are also provided in the dataset. Following Chen \textit{et al}. \cite{chen2012feature}, the first 800 frames are used to compose the training set and the remaining 1,200 are considered as the testing set. As shown in Table \ref{ucsd_mall}, we beat the second best approach by a 14.3\% improvement in MAE and 70\% improvement in MSE.

\noindent\textbf{WorldExpo'10.} The WorldExpo'10 dataset is introduced by Zhang \textit{et al}. \cite{zhang2015cross}, consisting of total 3980 frames extracted from 1132 video sequences captured with 108 surveillance equipment in Shanghai 2010 WorldExpo. Each sequence has 50 frame rates and a resolution of 576x720 pixels. For the test set, those frames are split into five scenes, named $S1 \sim S5$ respectively. Besides, the ROI (region of interest) and the perspective maps are provided for this dataset. Due to the abundant surveillance video data, this dataset is suitable to verify our model in visual surveillance. The comparison results of performance between our HSRNet with some previous methods are reported in Table \ref{world_table}. Overall, our model achieves the best average MAE performance compared with existing approaches.

\noindent\textbf{TRANCOS.} TRANCOS dataset \cite{guerrero2015extremely} is a vehicle counting dataset which consists of 1244 images of various traffic scenes captured by real video surveillance cameras with total 46796 annotated vehicles. Also, the ROI per image is provided. Images in the dataset have very different traffic scenarios without perspective maps. During the training period, 800 patches with the size of 115x115 pixels are randomly cropped from each image, and the ground truth density maps are generated via blurring annotations with 2D Gaussian Kernel. Different from counting dataset, we use the GAME for evaluation during the test period which can be formulated as follows:
\begin{equation}
    GAME(L)=\frac{1}{N}\sum_{n=1}^{N}(\sum_{l=1}^{4^{L}}|D_{I_{n}}^{l}-D_{I_{n}^{gt}}^{l}|)
\end{equation}
where $N$ denotes the amount of images. Given a specific number $L$, the GAME(L) divides each image into $4^{L}$ non-overlapping regions of equal area, $D_{I_{n}}^{l}$ is the estimated count for image $n$ within region $l$, and $D_{I_{n}^{gt}}^{l}$ is the corresponding ground truth count. Note that GAME(0) equals to the aforementioned MAE metric. Results shown in Table \ref{trancos_table} indicates that HSRNet achieves the state-of-art performance on four GAME metrics, which demonstrates the generalization ability of our model.

\begin{figure*}[t]
\centerline{\includegraphics[width=\linewidth]{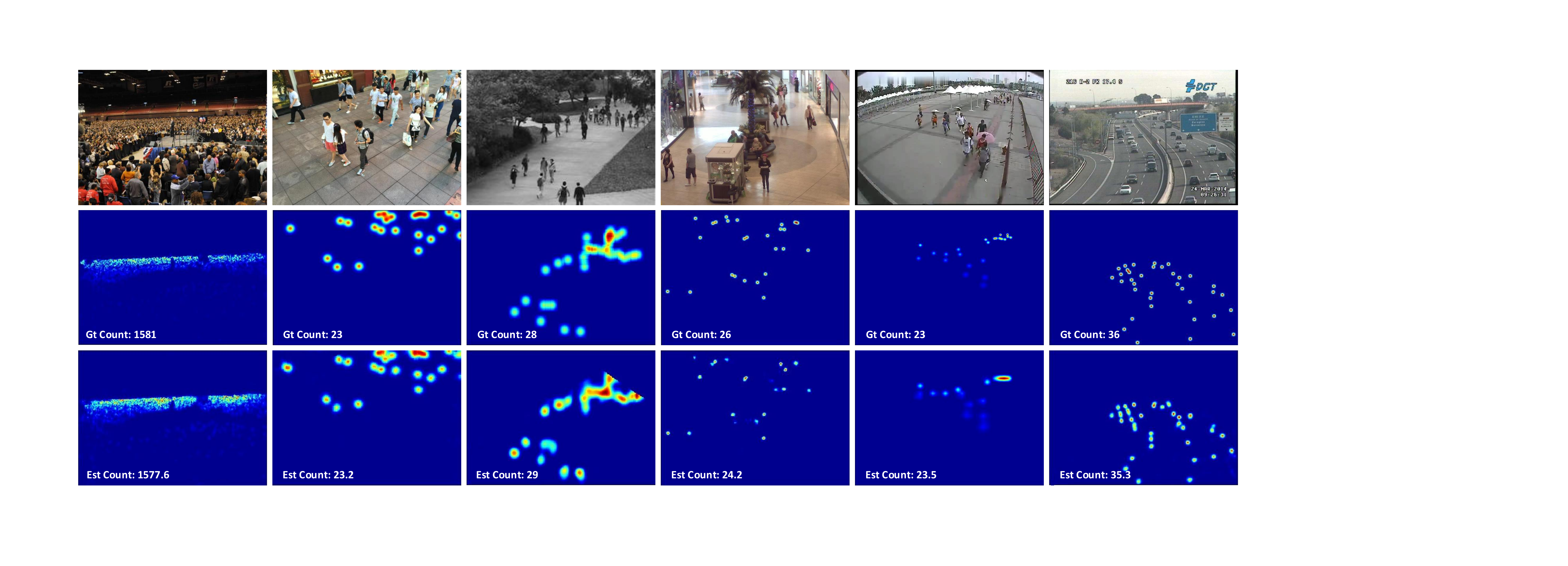}}
\caption{From left to right, images are taken from ShanghaiTech Part A, ShanghaiTech Part B, UCSD, MALL, WorldExpo'10, TRANCOS datasets. The second row shows the groundtruth density maps and the third row displays the output density maps generated by our HSRNet.} \label{visualization}
\end{figure*}

Qualitatively, we visualize the density maps generated by the proposed HSRNet on these five datasets in Figure \ref{visualization}. It is worth noting that our model can generate high-quality density maps and produce accurate crowd counts.  

\subsection{Ablation Study}
In this section, we conduct further experiments to explore the detail of the model design and network parameters. All experiments in this section are performed on ShanghaiTech dataset for its large scale variations.

\begin{table}
\renewcommand{\arraystretch}{1.2}
\caption{Experimental results on the TRANCOS dataset.}
\label{trancos_table}
\centering
\begin{tabular}{c|c|c|c|c}
\hline
 Method &  GAME 0 &  GAME 1 &  GAME 2 &  GAME 3  \\
\hline
\hline
Fiaschi \textit{et al}. \cite{fiaschi2012learning} & 17.77 & 20.14 & 23.65 & 25.99 \\
\hline
Lempitsky \textit{et al}. \cite{lempitsky2010learning}& 13.76 & 16.72 & 20.72 & 24.36 \\
\hline
Hydra-3s\cite{onoro2016towards} & 10.99 & 13.75 & 16.69 & 19.32 \\
\hline
FCN-HA \cite{zhang2017fcn}& 4.21 & - & - & - \\
\hline
CSRNet \cite{li2018csrnet}& 3.56 & 5.49 & 8.57 & 15.04 \\
\hline 
HSRNet (ours) & \textbf{3.03} & \textbf{4.57} & \textbf{6.46} & \textbf{9.68} \\
\hline
\end{tabular}
\end{table}

\noindent\textbf{Architecture learning.} We first evaluate the impact of each component in our architecture by separating all the modules and reorganizing them step by step. We perform this experiment on ShanghaiTech Part A dataset and the results are listed in Table \ref{architecture_design}. The backbone refers to the VGG-16 model. We add a $1 \times 1$ convolutional layer to the end to generate the density map, which is defined as our baseline. It is obvious that combining backbone with the Scale Recalibration Module (SRM) can boost the performance (MAE 75.6 vs 73.7), which verifies the effectiveness of the SRM module. We divide the Scale Focus Module (SFM) into two parts: Channel Focus (CF) and Spatial Focus (SF). The third row and the fourth row verify their respective significance (MAE 69.9,68.8 vs 73.7). Besides, the combination of the two parts is more effective than using one of them alone.
We add the Scale Consistency loss (SC) to supervise the model learning. This strategy also brings a significant improvement to the performance (MAE 65.1 vs 62.3). Overall, each part in the model is effective and complementary to each other, which can significantly boost the performance on the final results.

\begin{figure}[t]
\centerline{\includegraphics[width=\linewidth]{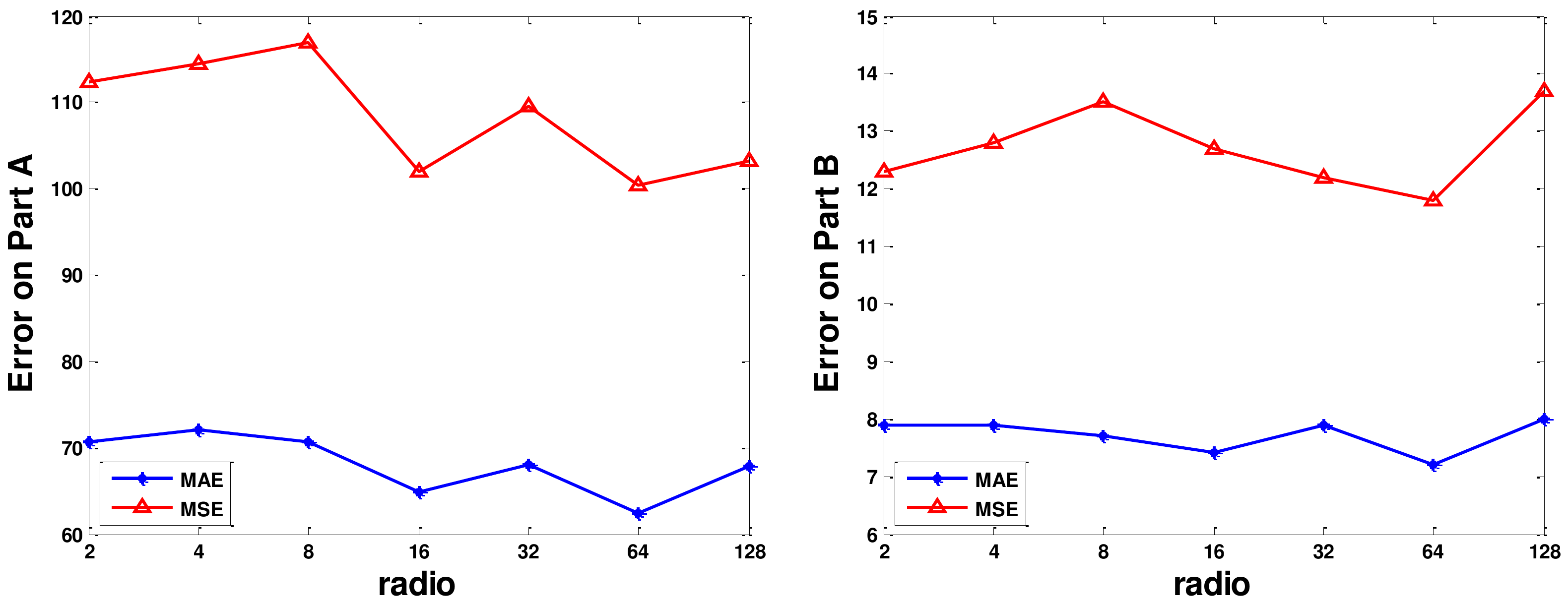}}
\caption{The line chart of performance under different ratios inside the Channel Focus on ShanghaiTech Part A and Part B dataset.} \label{ratio}
\end{figure}

\noindent\textbf{Ratio of Channel Focus.} We measure the performance of HSRNet with different ratios of Channel Focus introduced in Eqn. \ref{ratio_eq}. On the one hand, the ratio needs to be small enough to ensure the representative capability of full connections, on the other hand, if the ratio is too small, then parameters will become more numerous and may introduce computational redundancy. To find a balance between capability and computational cost, experiments are performed for a series of ratio values. Specifically, we gradually increase the ratio at twice the interval and results are shown in Figure \ref{ratio}. As the ratio increases, the error estimation undergoes a process of decreasing first and then increasing. The proposed model delivers the best accuracy on both Part A and Part B of ShanghaiTech dataset when the ratio equals to 64. Therefore, this value is used for all experiments in this paper.

\begin{table}
\renewcommand{\arraystretch}{1.2}
\caption{Experimental results of architecture learning on ShanghaiTech Part A dataset.}
\label{architecture_design}
\centering
\begin{tabular}{ccccc|cc}
\hline
 Backbone & SRM &  CF &  SF & SC & MAE &  MSE  \\
\hline
\hline
\checkmark & $\times$ & $\times$ & $\times$ & $\times$ & 75.6 & 118.7 \\

\checkmark & \checkmark & $\times$ & $\times$ & $\times$ & 73.7&  114.8\\

\checkmark & \checkmark & \checkmark & $\times$ &$\times$& 69.9 & 108.9 \\

\checkmark & \checkmark & $\times$ & \checkmark &$\times$& 68.8 & 107.2\\ 

\checkmark & \checkmark & \checkmark & \checkmark &$\times$& 65.1 &104.3\\

\checkmark & \checkmark & \checkmark & \checkmark & \checkmark & \textbf{62.3} & \textbf{100.3}\\
\hline
\end{tabular}
\end{table}

\begin{figure*}[t]
\centerline{\includegraphics[width=\linewidth]{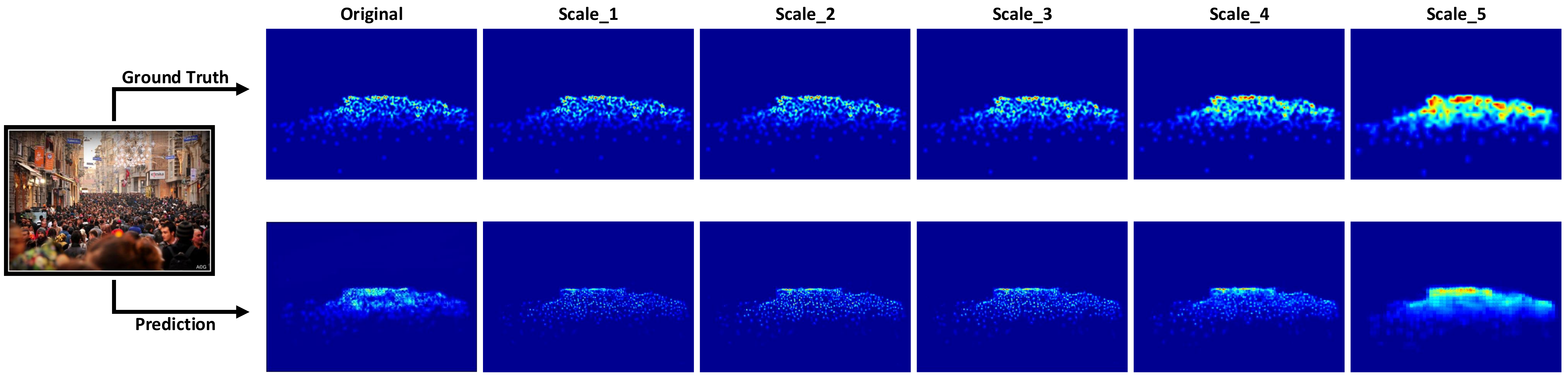}}
\caption{From left to right, the images pyramid contains the 1,1,1/2,1/4,1/8,1/16 size of the original image resolution. The first row indicates the groundtruth density maps while the second row is the density maps generated by the proposed HSRNet.} \label{scale}
\end{figure*}

\noindent\textbf{Sequence of SFM.} We evaluate the effect of the sequence of Channel Focus and Spatial Focus in the proposed Scale Focus Module (SFM). Therefore, we design four networks which are different from each other for the design of SFM module and experimental results are shown in Table \ref{sam_design}. Channel+Spatial refers to the network with Channel Focus module ahead Spatial Focus while Spatial+Channel refers to the opposite one. Apart from the serial settings, we also design a parallel one which feeds the input separately to the Channel focus and Spatial focus part inside the SFM. Then we have two choices: the one is average their output as the final output, the other is utilizing a convolution layer to process the stack of these two features. We name these two networks as (Channel $\oplus$ Spatial) + average and (Channel $\oplus$ Spatial) + Conv respectively. We conduct experiments on ShanghaiTech Part A and Part B datasets. The results mainly illustrates two aspects: 1) the stacked design of two modules are more effective than the parallel design, 2) Channel+Spatial is the most optimal choice to achieve the best accuracy, which demonstrates the validity of the model design.

\begin{table}
\renewcommand{\arraystretch}{1.2}
\caption{Validation of sequences inside the proposed Scale Focus Module.}
\label{sam_design}
\centering
\begin{tabular}{c|c|c|c|c}
\hline
 \multirow{2}{*}{\bfseries{Sequence inside the SFM}}&\multicolumn{2}{|c|}{\bfseries{Part\underline{\hbox to 0.1cm{}}A}} & \multicolumn{2}{|c}{\bfseries{
 Part\underline{\hbox to 0.1cm{}}B}}\\
\cline{2-5}
 & MAE &  MSE &  MAE &  MSE \\
\hline
Spatial+Channel & 64.9 & 108.3 &8.1 & 12.1\\
\hline
Channel+Spatial & \textbf{62.3} & \textbf{100.3} &\textbf{7.2} &\textbf{11.8}\\
\hline
(Channel $\oplus$ Spatial) + average &  68.8 & 107.2 & 11.0 & 19.8\\
\hline
(Channel $\oplus$ Spatial) + Conv & 66.2 & 104.3 & 10.2 & 18.9 \\
\hline
\end{tabular}
\end{table}

\noindent\textbf{Scale Consistency.} To understand the effect of the Scale Consistency loss more deeply, we visualization the immediate results of the proposed HSRNet and compare them with the groundtruth density map pyramid. As shown in Figure \ref{scale}, the $scale\_i$ represents the filter size of the average pooling operation on the groundtruth maps. Noted that the scale-associated outputs are closer to their corresponding groundtruth density maps. With the supervision of the extra Scale Consistency loss, the responses of the immediate stages in the network are indeed associated with the scales of pedestrians rather than stay the same with the groundtruth map. For instance, the shallow layers (such as $Scale_1,Scale_2$) are more sensitive to the small scale of pedestrians, while the deep layers (such as $Scale_4,Scale_5$) perform well on the large scale of crowds. By fusing these outputs, the final result can cover the multi-scale crowd distributions in complicated scenes.

\begin{figure}[t]
\centerline{\includegraphics[width=0.8\linewidth]{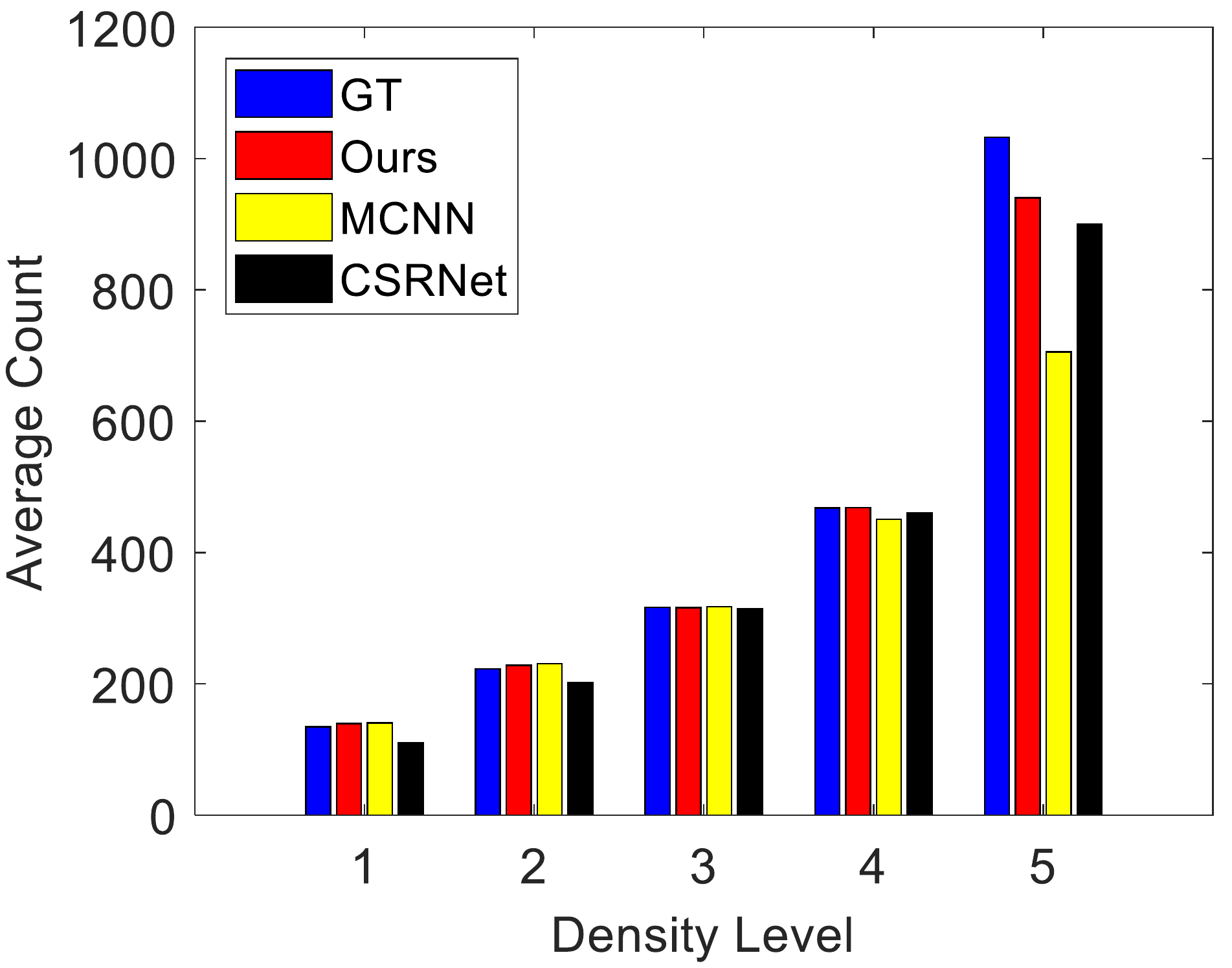}}
\caption{The histogram of average counts estimated by different approaches on five density levels from ShanghaiTech Part A dataset.} \label{density_level}
\end{figure}

\noindent\textbf{Scale Invariance.} We turn to evaluate the scale invariance of the feature representations from different stages in the proposed HSRNet for diverse scenes with various crowd counts. To achieve this, we divide the ShanghaiTech Part A test set into five groups according to the number of people in each scene. Each set represents a specific density level. The histogram of the results can be observed in Figure \ref{density_level}. The increase in density level represents an increase in the average number of people. We compare our method with two existing classic representative counting networks, MCNN \cite{7780439} and CSRNet \cite{li2018csrnet}. It is obvious that MCNN performs well on the relatively sparse scenes while loses its superiority on the dense crowds. The performance of the CSRNet tends to be the opposite. Noted that the proposed HSRNet outperforms the two models over all groups, which further demonstrates the scale generalization of our model on highly complicated scenes.

\section{CONCLUSION}
In this paper, we propose a single column but multi-scale network named as Hierarchical Scale Recalibration Network (HSRNet), which can exploit global contextual information and aggregate multi-scale information simultaneously. The proposed HSRNet consists of two main parts: Stacked Focus Module (SFM) and Scale Recalibration Module (SRM). Specifically, SFM models the global contextual dependencies among channel and spatial dimensions, which contributes to generating more informative feature representations. Besides, SRM recalibrates the feature responses generated by the SAM to generate multi-scale predictions, and then utilize a scale-specific fusion strategy to aggregate scale-associated outputs to generate the final density maps. Besides, we design a Scale Consistency loss to enhance the learning of scale-associated outputs towards their corresponding multi-scale groundtruth density maps. With the proposed modules combined, the network can tackle the difficulties of scale variations and generate more precise density maps in highly congested crowd scenes. Extensive experiments on four counting benchmark datasets and one vehicle dataset show that our method delivers the state-of-the-art performance over existing approaches and can be extended to other tasks. Besides, throughout ablation studies are conducted on ShanghaiTech dataset to validate the effectiveness of each part in the proposed HSRNet.

\bibliography{ecai}

\begin{thebibliography}{10}

\bibitem{babu2018divide}
Deepak Babu~Sam, Neeraj~N Sajjan, R~Venkatesh~Babu, and Mukundhan Srinivasan,
  `Divide and grow: Capturing huge diversity in crowd images with incrementally
  growing cnn', in {\em Proceedings of the IEEE Conference on Computer Vision
  and Pattern Recognition}, pp. 3618--3626, (2018).

\bibitem{bai2019crowd}
Haoyue Bai, Song Wen, and S-H Gary~Chan, `Crowd counting on images with scale
  variation and isolated clusters', in {\em Proceedings of the IEEE
  International Conference on Computer Vision Workshops}, pp. 0--0, (2019).

\bibitem{bartos2016learning}
Karel Bartos, Michal Sofka, and Vojtech Franc, `Learning invariant
  representation for malicious network traffic detection', in {\em Proceedings
  of the Twenty-second European Conference on Artificial Intelligence}, pp.
  1132--1139. IOS Press, (2016).

\bibitem{cao2018scale}
Xinkun Cao, Zhipeng Wang, Yanyun Zhao, and Fei Su, `Scale aggregation network
  for accurate and efficient crowd counting', in {\em Proceedings of the
  European Conference on Computer Vision (ECCV)}, pp. 734--750, (2018).

\bibitem{chan2008privacy}
Antoni~B Chan, Zhang-Sheng~John Liang, and Nuno Vasconcelos, `Privacy
  preserving crowd monitoring: Counting people without people models or
  tracking', in {\em 2008 IEEE Conference on Computer Vision and Pattern
  Recognition}, pp. 1--7. IEEE, (2008).

\bibitem{chen2012feature}
Ke~Chen, Chen~Change Loy, Shaogang Gong, and Tony Xiang, `Feature mining for
  localised crowd counting.', in {\em BMVC}, volume~1, p.~3, (2012).

\bibitem{cheng2019improving}
Zhi-Qi Cheng, Jun-Xiu Li, Qi~Dai, Xiao Wu, Jun-Yan He, and Alexander~G
  Hauptmann, `Improving the learning of multi-column convolutional neural
  network for crowd counting', in {\em Proceedings of the 27th ACM
  International Conference on Multimedia}, pp. 1897--1906. ACM, (2019).

\bibitem{fiaschi2012learning}
Luca Fiaschi, Ullrich K{\"o}the, Rahul Nair, and Fred~A Hamprecht, `Learning to
  count with regression forest and structured labels', in {\em Proceedings of
  the 21st International Conference on Pattern Recognition (ICPR2012)}, pp.
  2685--2688. IEEE, (2012).

\bibitem{french2015convolutional}
Geoffrey French, MH~Fisher, Michal Mackiewicz, and Coby Needle, `Convolutional
  neural networks for counting fish in fisheries surveillance video', {\em
  Proceedings of the machine vision of animals and their behaviour (MVAB)},
  7--1, (2015).

\bibitem{guerrero2015extremely}
Ricardo Guerrero-G{\'o}mez-Olmedo, Beatriz Torre-Jim{\'e}nez, Roberto
  L{\'o}pez-Sastre, Saturnino Maldonado-Basc{\'o}n, and Daniel Onoro-Rubio,
  `Extremely overlapping vehicle counting', in {\em Iberian Conference on
  Pattern Recognition and Image Analysis}, pp. 423--431. Springer, (2015).

\bibitem{8578843}
J.~{Hu}, L.~{Shen}, and G.~{Sun}, `Squeeze-and-excitation networks', in {\em
  2018 IEEE/CVF Conference on Computer Vision and Pattern Recognition}, pp.
  7132--7141, (June 2018).

\bibitem{jiang2019crowd}
Xiaolong Jiang, Zehao Xiao, Baochang Zhang, Xiantong Zhen, Xianbin Cao, David
  Doermann, and Ling Shao, `Crowd counting and density estimation by trellis
  encoder-decoder networks', in {\em Proceedings of the IEEE Conference on
  Computer Vision and Pattern Recognition}, pp. 6133--6142, (2019).

\bibitem{kang2018crowd}
Di~Kang and Antoni Chan, `Crowd counting by adaptively fusing predictions from
  an image pyramid', {\em arXiv preprint arXiv:1805.06115}, (2018).

\bibitem{kang2018beyond}
Di~Kang, Zheng Ma, and Antoni~B Chan, `Beyond counting: Comparisons of density
  maps for crowd analysis tasks—counting, detection, and tracking', {\em IEEE
  Transactions on Circuits and Systems for Video Technology}, {\bf 29}(5),
  1408--1422, (2018).

\bibitem{kingma2014adam}
Diederik~P Kingma and Jimmy Ba, `Adam: A method for stochastic optimization',
  {\em arXiv preprint arXiv:1412.6980}, (2014).

\bibitem{lempitsky2010learning}
Victor Lempitsky and Andrew Zisserman, `Learning to count objects in images',
  in {\em Advances in neural information processing systems}, pp. 1324--1332,
  (2010).

\bibitem{li2018csrnet}
Yuhong Li, Xiaofan Zhang, and Deming Chen, `Csrnet: Dilated convolutional
  neural networks for understanding the highly congested scenes', in {\em
  Proceedings of the IEEE conference on computer vision and pattern
  recognition}, pp. 1091--1100, (2018).

\bibitem{liu2018leveraging}
Xialei Liu, Joost van~de Weijer, and Andrew~D Bagdanov, `Leveraging unlabeled
  data for crowd counting by learning to rank', in {\em Proceedings of the IEEE
  Conference on Computer Vision and Pattern Recognition}, pp. 7661--7669,
  (2018).

\bibitem{onoro2016towards}
Daniel Onoro-Rubio and Roberto~J L{\'o}pez-Sastre, `Towards perspective-free
  object counting with deep learning', in {\em European Conference on Computer
  Vision}, pp. 615--629. Springer, (2016).

\bibitem{sam2017switching}
Deepak~Babu Sam, Shiv Surya, and R~Venkatesh Babu, `Switching convolutional
  neural network for crowd counting', in {\em 2017 IEEE Conference on Computer
  Vision and Pattern Recognition (CVPR)}, pp. 4031--4039. IEEE, (2017).

\bibitem{simonyan2014very}
Karen Simonyan and Andrew Zisserman, `Very deep convolutional networks for
  large-scale image recognition', {\em arXiv preprint arXiv:1409.1556}, (2014).

\bibitem{sindagi2017cnn}
Vishwanath~A Sindagi and Vishal~M Patel, `Cnn-based cascaded multi-task
  learning of high-level prior and density estimation for crowd counting', in
  {\em 2017 14th IEEE International Conference on Advanced Video and Signal
  Based Surveillance (AVSS)}, pp. 1--6. IEEE, (2017).

\bibitem{sindagi2017generating}
Vishwanath~A Sindagi and Vishal~M Patel, `Generating high-quality crowd density
  maps using contextual pyramid cnns', in {\em Proceedings of the IEEE
  International Conference on Computer Vision}, pp. 1861--1870, (2017).

\bibitem{walach2016learning}
Elad Walach and Lior Wolf, `Learning to count with cnn boosting', in {\em
  European Conference on Computer Vision}, pp. 660--676. Springer, (2016).

\bibitem{wang2019learning}
Qi~Wang, Junyu Gao, Wei Lin, and Yuan Yuan, `Learning from synthetic data for
  crowd counting in the wild', in {\em Proceedings of the IEEE Conference on
  Computer Vision and Pattern Recognition}, pp. 8198--8207, (2019).

\bibitem{woo2018cbam}
Sanghyun Woo, Jongchan Park, Joon-Young Lee, and In~So~Kweon, `Cbam:
  Convolutional block attention module', in {\em Proceedings of the European
  Conference on Computer Vision (ECCV)}, pp. 3--19, (2018).

\bibitem{wu2005detection}
Bo~Wu and Ramakant Nevatia, `Detection of multiple, partially occluded humans
  in a single image by bayesian combination of edgelet part detectors', in {\em
  Tenth IEEE International Conference on Computer Vision (ICCV'05) Volume 1},
  volume~1, pp. 90--97. IEEE, (2005).

\bibitem{xiong2017spatiotemporal}
Feng Xiong, Xingjian Shi, and Dit-Yan Yeung, `Spatiotemporal modeling for crowd
  counting in videos', in {\em Proceedings of the IEEE International Conference
  on Computer Vision}, pp. 5151--5159, (2017).

\bibitem{xiong2016person}
Mingfu Xiong, Jun Chen, Zheng Wang, Zhongyuan Wang, Ruimin Hu, Chao Liang, and
  Daming Shi, `Person re-identification via multiple coarse-to-fine deep
  metrics', in {\em Proceedings of the Twenty-second European Conference on
  Artificial Intelligence}, pp. 355--362. IOS Press, (2016).

\bibitem{zhang2019attentional}
Anran Zhang, Lei Yue, Jiayi Shen, Fan Zhu, Xiantong Zhen, Xianbin Cao, and Ling
  Shao, `Attentional neural fields for crowd counting', in {\em Proceedings of
  the IEEE International Conference on Computer Vision}, pp. 5714--5723,
  (2019).

\bibitem{zhang2015cross}
Cong Zhang, Hongsheng Li, Xiaogang Wang, and Xiaokang Yang, `Cross-scene crowd
  counting via deep convolutional neural networks', in {\em Proceedings of the
  IEEE conference on computer vision and pattern recognition}, pp. 833--841,
  (2015).

\bibitem{zhang2018crowd}
Lu~Zhang, Miaojing Shi, and Qiaobo Chen, `Crowd counting via scale-adaptive
  convolutional neural network', in {\em 2018 IEEE Winter Conference on
  Applications of Computer Vision (WACV)}, pp. 1113--1121. IEEE, (2018).

\bibitem{zhang2017fcn}
Shanghang Zhang, Guanhang Wu, Joao~P Costeira, and Jos{\'e}~MF Moura,
  `Fcn-rlstm: Deep spatio-temporal neural networks for vehicle counting in city
  cameras', in {\em Proceedings of the IEEE International Conference on
  Computer Vision}, pp. 3667--3676, (2017).

\bibitem{7780439}
Y.~{Zhang}, D.~{Zhou}, S.~{Chen}, S.~{Gao}, and Y.~{Ma}, `Single-image crowd
  counting via multi-column convolutional neural network', in {\em 2016 IEEE
  Conference on Computer Vision and Pattern Recognition (CVPR)}, pp. 589--597,
  (June 2016).

\bibitem{8497050}
Z.~{Zou}, X.~{Su}, X.~{Qu}, and P.~{Zhou}, `Da-net: Learning the fine-grained
  density distribution with deformation aggregation network', {\em IEEE
  Access}, {\bf 6},  60745--60756, (2018).

\bibitem{zou2019attend}
Zhikang Zou, Yu~Cheng, Xiaoye Qu, Shouling Ji, Xiaoxiao Guo, and Pan Zhou,
  `Attend to count: Crowd counting with adaptive capacity multi-scale cnns',
  {\em Neurocomputing}, {\bf 367},  75--83, (2019).

\bibitem{zou2019enhanced}
Zhikang Zou, Huiliang Shao, Xiaoye Qu, Wei Wei, and Pan Zhou, `Enhanced 3d
  convolutional networks for crowd counting', {\em arXiv preprint
  arXiv:1908.04121}, (2019).

\end{thebibliography}
\end{document}